# AlphaGAN: Generative adversarial networks for natural image matting

Sebastian Lutz
lutzs@scss.tcd.ie

Konstantinos Amplianitis
ampliank@scss.tcd.ie

Aljoša Smolić
smolica@scss.tcd.ie

V-SENSE
School of Computer Science and Statistics
Trinity College Dublin

## Abstract

We present the first generative adversarial network (GAN) for natural image matting. Our novel generator network is trained to predict visually appealing alphas with the addition of the adversarial loss from the discriminator that is trained to classify well-composed images. Further, we improve existing encoder-decoder architectures to better deal with the spatial localization issues inherited in convolutional neural networks (CNN) by using dilated convolutions to capture global context information without downscaling feature maps and losing spatial information. We present state-of-the-art results on the alphamatting online benchmark for the gradient error and give comparable results in others. Our method is particularly well suited for fine structures like hair, which is of great importance in practical matting applications, e.g. in film/TV production.

## 1 Introduction

Natural image matting is defined as the problem of accurately estimating the opacity of a foreground object in an image or video sequence. It is a field that has received significant attention from the scientific community as it is used in many image-editing and film post-production applications. With the recent advances in mobile technology, high-quality matting algorithms are required for compositing tasks, both for professional and ordinary users. Formally, image matting approaches require as input an image, which is expected to contain a foreground object and the image background. Mathematically, every pixel $i$ in the image is assumed to be a linear combination of the foreground and background colors, expressed as:

$$I_i = \alpha_i F_i + (1 - \alpha_i) B_i, \quad \alpha_i \in [0, 1] \tag{1}$$

where $\alpha_i$ is a scalar value that defines the foreground opacity at pixel $i$ and is referred to as the alpha value. Since neither the foreground, nor the background RGB values are known, this is a severely ill-posed problem, consisting of 7 unknown and only 3 known values. Typically, some additional information in the form of *scribbles* [31] or a *trimap* [8] is given as additional information to decrease the difficulty of the problem. Both additional input methods already roughly segment the image in foreground, background and regions with unknown opacity. Generally they serve as initialization information and many methods propagate the alpha values from known image regions to the unknown region.

Most current algorithms aim to solve the matting equation 1 in a closed-form manner and treat it as a color-problem by either sampling or affinity-based methods. This over-dependency solely on color information can lead to artifacts in images where the foreground and background color distributions overlap, which is often the case in natural images [33].

Most state-of-the-art algorithms in other computer vision tasks nowadays rely on deep convolutional neural networks, which are able to learn structural information and abstract representations of images. Until recently, this was impossible for natural image matting since a large amount of training data is needed to train CNNs, which wasn't available back then. However, Xu et al. [33] released a new matting dataset and have shown that it can be used to train CNNs for natural image matting and reach state-of-the-art performance on the alphamatting.com [25] dataset. However, this dataset only consists of 431 unique foreground objects with corresponding alpha ground-truth and the large dataset size is only reached by compositing a large amount of new images using random backgrounds.

Our approach builds on the CNN by Xu et al. [33] and improves it in several ways to reach state-of-the-art performance on the natural image matting benchmark [25].

**Our Contribution.** We propose a generative adversarial network (GAN) for natural image matting. We improve on the network architecture of Xu et al. [33] to better deal with the spatial localization issues inherent in CNNs by using dilated convolutions to capture global context information without downscaling feature maps and losing spatial information. Furthermore, we improve on the decoder structure of the network and use it as the generator in our generative adversarial model. The discriminator is trained on images that have been composited with the ground-truth alpha and the predicted alpha and therefore learns to recognize images that have been composited well, which helps the generator learn alpha predictions that lead to visually appealing compositions.

## 2    Previous Work

In this section, we briefly review traditional approaches for natural image matting, as well as more recent approaches using deep learning.

### 2.1    Local sample-based natural image matting

A significant amount of literature has been introduced over the last years for solving the ill-posed problem of natural image matting. The motivation behind these approaches is that they use color (sometimes also position) of user-defined foreground and background samples to infer the alpha values of the unknown regions in the image. Existing methods follow a *sampling* or *propagation* approach. In sample-based approaches, the known foreground and background samples that are in the near vicinity of the unknown pixel in question, should be also very "close" to the true foreground and background colors of that pixel and thus should be further processed to estimate the corresponding alpha value based on Eq. 1. However, it should be stressed that the meaning of "near" in this context is something very vague and existing methods deal with this problem in different ways. Bayesian matting [8], iterative matting [31], shared sampling matting [10], [13] and more recent approaches such as sparse coding [9] are some of the methods that follow this assumption.

Propagation approaches work by propagating the known alpha value between known local foreground and background samples to the unknown pixels. Approaches such as the Poisson matting [30], random walk [12], geodesic matting [2], spectral matting [20], close-form matting [21] and fuzzy connectedness matting [34] are some of the most known propagation methods introduced in this direction. The manifold preserving edit propagation algorithm [6] and information-flow matting [1] are more recent approaches. A detailed description on the above methods can be found in the survey work of Want et al. [32] as this analysis goes beyond the scope of our work.

## 2.2 Deep learning in natural image matting

Recently, a few deep learning methods were introduced for natural image matting. Specifically, Xu et al. [33] proposed a two-stage network, consisting of an encoder-decoder stage and a refinement stage. The first stage takes an image and the corresponding trimap as an input and predicts the alpha matte of the unknown trimap region. The output of the first stage is then given as an input to a small convolutional neural network that refines the alpha values and sharpens the edges. Shen et al. [28] proposed a fully automatic matting system for portrait photos based on an end-to-end convolutional neural network. A portrait image is given as an input along with a pre-trained shape mask which is used for automatically generating a trimap region. The alpha values of the trimap area are then computed from the proposed CNN. Furthermore, Cho et al. [7] proposed an end-to-end CNN architecture that utilizes the results deduced from local (closed-form matting [21]) and non-local (KNN matting [5]) matting algorithms along with RGB color images and learns the mapping between the input images and the reconstructed alpha mattes. Hu et al., [16] proposed a granular deep learning (GDL) architecture for the task of foreground-background separation. In their approach they created a hierarchical structure of a layered neural network designed as a granular system.
To the best of our knowledge, this work is the first approach using generative adversarial neural networks for natural image matting. However, GANs have shown good performance in other computer vision tasks, such as image-to-image translation [18] [37], image generation [24] or image editing [36].

## 3 Our Approach

To tackle the problem of image matting, we use a generative adversarial network. The generator of this network is a convolutional encoder-decoder network that is trained both with help of the ground-truth alphas as well as the adversarial loss from the discriminator. We detail our network in more detail in the following sections.

### 3.1 Training dataset

Deep learning approaches need a lot of data to generalize well. Large datasets like Imagenet [26] and MSCOCO [23] have helped tremendously in this regard for several computer vision tasks. One of the problems of natural image matting, however, is that it is significantly more difficult to collect ground-truth data than for most other tasks. The quality of the ground-truth also needs to be very high, because the methods need to capture very fine differences in the alpha to provide good results. Thankfully a new matting dataset [33] consisting of 431 unique foreground objects and their corresponding alpha has recently been published. This dataset has finally made it possible to train deep networks such as ours. Nevertheless, 431 images is not enough to train on their own, so we enhance the dataset in the following way, similar to how Xu et al. [33] propose in their approach:

For every foreground object, a random background image from MSCOCO is taken, which allows us to composite a new unique image out of the foreground, the provided ground-truth alpha and the background image. For further data augmentation, we randomly rotate the foreground and alpha by *n* degrees, sampled from a normal distribution with a mean of 0 and standard deviation of 5. We then generate a trimap by dilating the ground-truth alpha with random kernel sizes from 2 to 20. Next, we randomly crop a rectangular part of the foreground, alpha, trimap and background images, centered on some pixel within the unknown region of the trimap of a size chosen randomly from $320 \times 320$ to $720 \times 720$, and resize it to $320 \times 320$. This allows the network to be more scale invariant. Finally, we randomly flip the cropped images to get the final foreground, alpha, trimap and background images, which will be used to composite a new image as part of the training process.

## 3.2 Network architecture

Xu et al. [33] have recently shown that it is possible to train an encoder-decoder network with their matting dataset to produce state-of-the-art results. We build on their approach and trained a deep generative adversarial network on the same dataset. Our AlphaGAN architecture consists of one generator *G* and one discriminator *D*. *G* takes an image composited from the foreground, alpha and a random background appended with the trimap as 4th-channel as input and attempts to predict the correct alpha. *D* tries to distinguish between *real* 4-channel inputs and *fake* inputs where the first 3 channels are composited from the foreground, background and the predicted alpha. The full objective of this network can be seen in 3.3.

### 3.2.1 Generator

Our generator consists of a an encoder-decoder network similar to those that have achieved good results in other computer vision tasks, such as semantic segmentation [4] [15]. For the encoder, we take the Resnet50 [14] architecture, pretrained on Imagenet [26] and convert the convolutions in the 3rd and 4th Resnet blocks to dilated convolutions with rate 2 and 4 respectively for a final output stride of 8, similar to Chen et al. [3]. Since the training inputs are fixed to a size of $320 \times 320$, this leads to a feature map size of $40 \times 40$ in the final feature map of Resnet block 4. Even though the feature maps are downsampled less often, the dilated convolutions can still capture the same global context of the original Resnet50 classification network, while not losing as much spatial information. After the Resnet block 4, we add the atrous spatial pyramid pooling (ASPP) module from [3] to resample features at several scales for accurately and efficiently predicting regions of an arbitrary scale. We then feed the output of the ASPP to the decoder part of the network. We also change the first layer of the network slightly to accommodate our 4-channel input by initializing the extra channel in the convolution layer with zeros.

The decoder part of the network is kept simple and consists of several convolution layers and skip connections from the encoder to improve the alpha prediction by reusing local information to capture fine structures in the image [18]. First, the output of the encoder is bilinearly upsampled 2 times so that the feature maps have the same spatial resolution as those coming from Resnet block 1, which have an output stride of 4. The final feature map from block 1 is fed into a $1 \times 1$ convolution layer to reduce the number of dimensions and then concatenated with the upsampled feature maps from the encoder. This is followed by three $3 \times 3$ convolutions that steadily reduce the number of dimensions to 64. The saved pooling indices from the max-pooling layer in the encoder are used to upsample these feature maps to an output stride of 2, where they are concatenated again with the feature maps of the same resolution from the encoder, followed by some convolution layers. Finally, the feature maps are upsampled again using fractionally-strided convolutions, concatenated with the RGB input image and fed to a final set of convolution layers. All of these layers are followed by ReLU activation functions and batch-normalization layers [17], except the last one, which is followed by the sigmoid activation function to scale the output of the generator between 0 and 1, as needed for an alpha prediction (See Figure 1). A table detailing all layers in the network can be seen in the supplementary material.

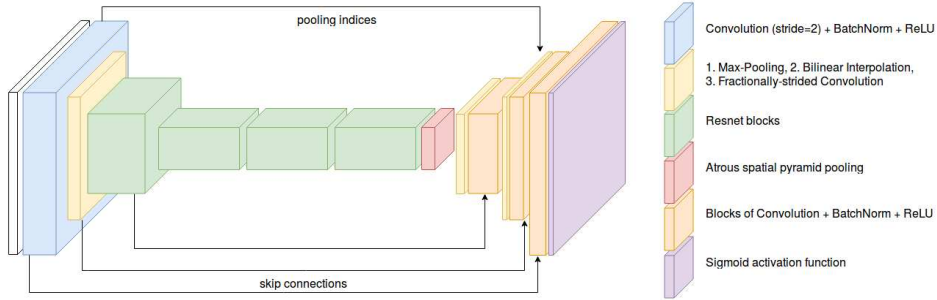

Figure 1: The generator is an encoder-decoder network with skip connections.

### 3.2.2 Discriminator

For the discriminator in our network, we use the PatchGAN introduced by Isola et al. [18]. This discriminator attempts to classify every $N \times N$ patch of the input as real or fake. The discriminator is run convolutionally over the input and all responses are averaged to calculate the final prediction of the discriminator $D$.

PatchGAN was designed to capture high-frequency structures and assumes independence between pixels that cannot be located in the same $N \times N$ patch. This suits the problem of alpha prediction, since the results of the generator trained only on the alpha-prediction loss can be overly smooth, as noted in [33]. The discriminator helps to alleviate this problem by forcing the generator to output sharper results. To help the discriminator focus on the right areas of the input and to guide the generator to predict alphas that would result in good compositions, the input of $D$ consists of 4 channels. The first 3-channels consist of the RGB values of a newly composited image, using the ground-truth foreground, a random background and the predicted alpha. The 4th channel is the input trimap to help guide the discriminator to focus on salient regions in the image. We found that for our network $N = 70$ is sufficient to balance good results and a low amount of parameters and running time of $D$.

## 3.3 Network objectives

The goal of our networks is to predict the true alpha of an image, given the trimap. In their paper Xu et al. [33] introduce two loss functions specifically for the problem of alpha matting, the alpha-prediction loss $\mathcal{L}_{alpha}$ and the compositional loss $\mathcal{L}_{comp}$. Additionally to these, we also use the adversarial loss [11] $\mathcal{L}_{GAN}$, which is defined as:

$$\mathcal{L}_{GAN}(G,D) = \log D(x) + \log \left(1 - D(C(G(x)))\right) \tag{2}$$

where $x$ is a *real* input: an image composited from the ground-truth alpha and foreground appended with the trimap. $C(y)$ is a composition function that takes the predicted alpha from $G$ as an input and uses it to composite a *fake* image. $G$ tries to generate alphas that are close to the ground-truth alpha, while $D$ tries to distinguish *real* from *fake* composited images. $G$ therefore tries to minimize $L_{GAN}$ against the discriminator $D$, which tries to maximize it.

The above losses are combined and lead to the full objective of our network:

$$\mathcal{L}_{AlphaGAN}(G,D) = \mathcal{L}_{alpha}(G) + \mathcal{L}_{comp}(G) + \mathcal{L}_{GAN}(G,D) \tag{3}$$

where we aim to solve $\arg\min_G \max_D \mathcal{L}_{AlphaGAN}$.

## 4 Experimental Results

In this section, we evaluate our approach on two datasets. The first one is the well-known alphamatting.com [25] evaluation benchmark, which consists of 28 training images and 8 test images. For each set, three different sizes of trimaps are provided, namely, "small", "large" and "user". The second one is the Composition-1k dataset [33], which includes 1000 test images composed from 50 unique foreground objects. We evaluate the quality of our results using the well known sum of absolute differences (SAD) and mean square error (MSE) but also the gradient and connectivity errors, which measure the matting quality as perceived by the human eye [25]. To avoid deviations from the original formulation of the metrics, as seen in other works ([35], [28]), we make use of the publicly available evaluation code provided by [33]. We use the default values for the gradient and connectivity error as proposed by the original authors of the evaluation metrics [25] throughout all our experiments.

### 4.1 Evaluating the network architecture

The Composition-1k test dataset consists of 1000 images composited out of 50 unique objects. However, since random background images are chosen when compositing, the resulting images do not look realistic, in the sense that they show scenes that do not exist in nature, e.g. a glass in the foreground floating before a woodland scene as the background. Further, the foreground and background images also have different characteristics, like lighting, noise, etc., that lead to images that cannot be considered natural.

Therefore, we mainly used the Composition-1k dataset to test our network architecture. We started by using a similar encoder-decoder architecture as [33], but replaced VGG16 [29] as encoder for Resnet50 [14]. We also tried other Resnet architectures, but found that Resnet50 performed best. By including the atrous pyramid pooling module (ASPP) [3] and using dilated convolutions for an output stride of 8 for the encoder, we further improved our performance. We also tried a multi-grid approach following [15], but found that this did not lead to better results. Finally, we added skip connections from the 1st and 2nd Resnet blocks and the RGB input to get to the final model we use for the generator. A comparison of some of the different network architectures that we tried are shown in Table 1.

| ASPP | OS=8 | OS=16 | MG | Skip | GAN loss | MSE |
|------|------|-------|-----|------|----------|------|
|  |  |  |  |  | ✓ | 0.049 |
| ✓ | ✓ |  |  |  | ✓ | 0.033 |
| ✓ | ✓ | ✓ |  |  | ✓ | 0.038 |
| ✓ | ✓ |  | ✓ |  | ✓ | 0.039 |
| ✓ | ✓ |  |  | ✓ | ✓ | **0.031** |
| ✓ | ✓ |  |  | ✓ |  | 0.041 |

Table 1: Comparison of different generator architectures. **ASPP**: Atrous spatial pyramid pooling [3], **OS**: Output stride of the final feature map, **MG**: Multi-Grid for dilated convolutions [15], **Skip**: Skip connections from the encoder, **GAN loss**: Additional adversarial loss during training.

| Method | SAD | MSE | Gradient ($\times 10^4$) | Connectivity ($\times 10^4$) |
|--------|-----|-----|---------------------------|-------------------------------|
| Shared Matting (SM) [10] | 117.0 (68.7) | 0.067 (0.032) | 10.1 (5.1) | 5.4 (5.2) |
| Comprehensive Sampling (CM) [27] | 56.5 (53.7) | 0.032 (0.030) | **3.4** (4.0) | 5.7 (5.4) |
| KNN Matting (KNN) [5] | 99.0 (53.6) | 0.070 (0.030) | 6.2 (4.0) | 8.5 (5.4) |
| DCNN Matting (DCNN) [7] | 155.8 (68.8) | 0.083 (0.032) | 11.5 (5.1) | 7.3 (6.0) |
| Three-layer Graph (TLGM) [22] | 106.4 (52.4) | 0.066 (0.030) | 7.0 (3.9) | 5.0 (4.3) |
| Information-flow Matting (IF) [1] | 75.4 (52.4) | 0.066 (0.030) | 6.3 (3.8) | 7.5 (5.3) |

Table 2: Quantitative results on the Composition-1k dataset. Our results are shown in parenthesis. We achieve better results than all the tested methods, with the sole exception marked in bold.

Additionally, we compare several top matting methods where there is public code available with our approach on the Composition-1k dataset [33]. For all methods, the original code from the authors is used, without any modifications. It was found that there were multiple failed cases when directly applied to the entire dataset. We believe that this is due to the inherently unrealistic nature of the dataset (see supplementary material for examples). To overcome this issue, we only provide comparison for results in which the images successfully produced a valid matting prediction. In contrast, our method succeeded in all image in the dataset. Quantitative results under all metrics are shown in Table 2. Our method delivers noticeably better results than the other approaches. The gradient error from the comprehensive sampling approach [27] is the only case where we do not achieve the best result as shown in Table 2. Some comparisons of results for this dataset can be seen in Figure 2. Additional results are provided in the supplementary material.

## 4.2 The alphamatting.com dataset

We submitted our results on the alphamatting.com benchmark [25] achieving state-of-the-art results for the Troll and Doll images, both for the SAD and MSE evaluation metrics and first place overall on the gradient evaluation metric. Even though we are not first in the SAD or MSE, our results are numerically very close to the top-performing results for the remaining images as shown in Table 3.

Overall, we achieve very visually appealing results, as seen in Figure 4 and by our results in the gradient metric which was introduced in [25] as a perceptually-friendly measure that had high correlation to good alpha mattes as perceived by humans. Similar to [19] we do not report the connectivity measure since it is not robust [25].

Our best results are for the Troll and Doll images, which is due to the ability of our approach to correctly predict the alpha values for very fine structures, like the hair. This is where the adversarial loss from the discriminator helps, since the discriminator is able to capture high-frequency structures and can distinguish between overly smooth predictions and ground-truth compositions during training, which allows the generator to learn to predict sharper structures. Our worst results come from the Net image. However, even though we appear low in the rankings for this image, we believe that our results still look very close to the top-performing approaches. Some examples of the alphamatting results are shown in Figure 4.

| | Average Rank* | | | | Troll | | | Doll | | | Donkey | | | Elephant | | | Plant | | | Pineapple | | | Plastic Bag | | | Net | | |
|---|---|---|---|---|---|---|---|---|---|---|---|---|---|---|---|---|---|---|---|---|---|---|---|---|---|---|---|---|
| | Overall | S | L | U | S | L | U | S | L | U | S | L | U | S | L | U | S | L | U | S | L | U | S | L | U | S | L | U |
| | | | | | | | | | | **Sum of Absolute Differences** | | | | | | | | | | | | | | | | | | |
| DI [33] | 4.6 | 5.6 | 3.6 | 4.6 | 10.7 | 11.2 | 11.0 | 4.8 | 5.8 | 5.6 | **2.8** | **2.9** | **2.9** | 1.1 | **1.1** | 2.0 | 6.0 | 7.1 | 8.9 | 2.7 | **3.2** | 3.9 | 19.2 | 19.6 | 18.7 | 21.8 | 23.9 | 24.1 |
| IF [11] | 5.4 | 6.5 | 4.9 | 4.8 | 10.3 | 11.2 | 12.5 | 5.6 | 7.3 | 7.3 | 3.8 | 4.1 | 3.0 | 1.4 | 2.3 | 2.0 | 5.9 | 7.1 | 8.6 | 3.6 | 5.7 | 4.6 | 18.3 | 19.3 | 15.8 | 20.2 | 22.2 | 22.3 |
| DCNN [7] | 6.8 | 8.6 | 4.9 | 7.0 | 12.0 | 14.1 | 14.5 | 5.3 | 6.4 | 6.8 | 3.9 | 4.5 | 3.4 | 1.6 | 2.5 | 2.2 | 6.0 | 6.9 | 9.1 | 4.0 | 6.0 | 5.3 | 19.9 | 19.2 | 19.1 | 19.4 | 20.0 | 21.2 |
| Ours | 7.8 | 8.6 | 7.5 | 7.4 | **9.6** | **10.7** | **10.4** | **4.7** | **5.3** | **5.4** | 3.1 | 3.7 | 3.1 | 1.3 | 2.0 | 2.0 | 6.4 | 8.3 | 9.3 | 3.6 | 5.0 | 4.3 | 20.8 | 21.5 | 20.6 | 25.7 | 26.7 | 26.7 |
| TLGM [22] | 11.5 | 8.1 | 8.9 | 17.6 | 10.7 | 15.2 | 13.8 | 4.9 | 5.6 | 8.1 | 3.9 | 4.4 | 3.6 | **1.0** | 1.8 | 3.0 | 5.9 | 7.3 | 12.4 | 4.2 | 8.0 | 8.5 | 24.2 | 25.6 | 24.2 | 20.5 | 23.5 | 22.2 |
| | | | | | | | | | | **Gradient** | | | | | | | | | | | | | | | | | | |
| Ours | 9.3 | 8.0 | 6.8 | 13.3 | 0.2 | 0.2 | 0.2 | **0.2** | 0.3 | 0.2 | 0.2 | 0.3 | 0.3 | 0.2 | **0.2** | 0.4 | 1.8 | 2.4 | 2.7 | 1.1 | 1.5 | 1.5 | 0.9 | 1.1 | 1.0 | 0.5 | 0.5 | 0.6 |
| DCNN [7] | 10.9 | 13.6 | 10.4 | 8.8 | 0.2 | 0.2 | 0.2 | 0.2 | 0.3 | 0.3 | 0.3 | 0.4 | 0.3 | 0.3 | 0.4 | 1.5 | 2.5 | 2.1 | 1.1 | 1.5 | 1.4 | 1.0 | 0.6 | 0.6 | 0.5 |
| DI [33] | 11.4 | 8.1 | 8.4 | 17.6 | 0.4 | 0.4 | 0.5 | **0.2** | **0.1** | **0.1** | 0.2 | 0.6 | 1.3 | 1.5 | 2.4 | 0.8 | 0.9 | 1.3 | **0.7** | **0.8** | 0.1 | 0.4 | 0.3 | 0.5 | 0.5 | 0.5 |
| IF [11] | 12.5 | 15.1 | 10.1 | 12.1 | 0.2 | **0.2** | 0.2 | 0.2 | 0.4 | 0.4 | 0.4 | 0.3 | 0.3 | 0.4 | 0.4 | 1.7 | 1.8 | 2.2 | 0.9 | 1.3 | 1.3 | 1.5 | 1.4 | 0.8 | 0.5 | 0.6 | 0.5 |
| TLGM [22] | 14.6 | 11.6 | 11.8 | 20.5 | **0.2** | 0.2 | 0.2 | 0.2 | 0.4 | 0.4 | 0.3 | 0.3 | **0.1** | 0.4 | 1.7 | 2.7 | 1.1 | 1.9 | 2.4 | 1.6 | 1.0 | 0.5 | 0.6 | 0.4 |

Table 3: SAD and gradient results for the top five methods on the alphamatting.com dataset. Best results are shown in bold.

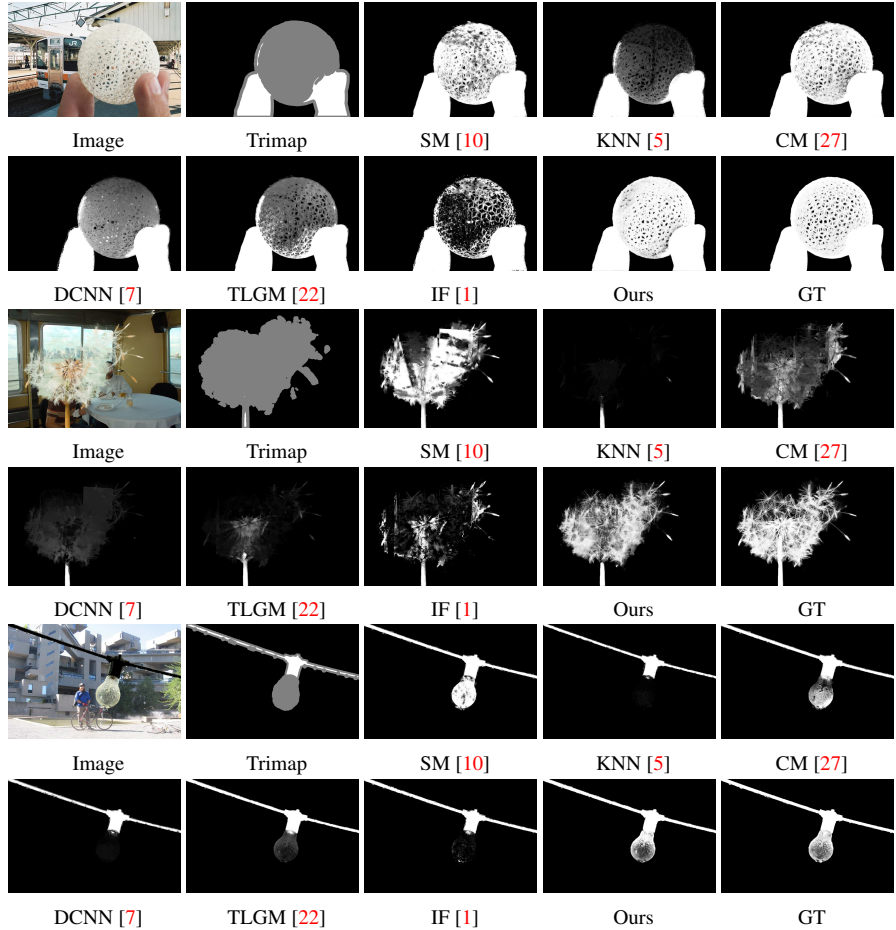

Figure 2: Comparison of results on the Composition-1k testing dataset.

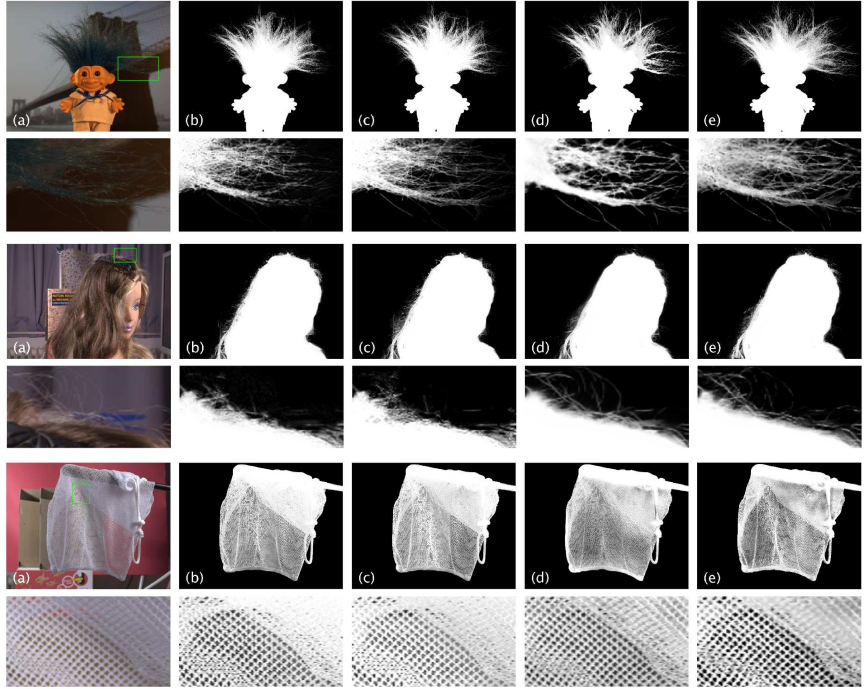

Figure 3: Alpha matting predictions for the "Troll" and "Doll" images (our best results) and the "Net" image (our worst result) taken from the alphamatting.com dataset. From left to right: DCNN [7], IF [1], DI [33], Ours.

## 5 Conclusion

In this paper we proposed a novel generative adversarial network architecture for the problem of natural image matting. To the best of our knowledge, this is the first work that uses GANs for this computer vision task. Our generator is trained to predict alpha mattes from input images while the discriminator is trained to distinguish good images composited from the ground-truth alpha from images composited with the predicted alpha. Additionally, we introduce some network enhancements to the generator that have been shown to give an increase in performance for the task of semantic segmentation. These changes allow us to train the network to predict alphas that lead to visually appealing compositions, as our results in the alphamatting benchmark show. Our method ranks first in this benchmark for the gradient metric, which was designed as a perceptual measure. For all the other metrics we show comparable results to the state-of-the-art and are first in the SAD and MSE errors for the Troll and Doll images. Our results in these images especially manage to capture the high-frequency hair structures, which might be attributed to the addition of the adversarial loss during training. Additionally, we compare with publicly available methods on the Composition-1k test dataset and achieve state-of-the-art results.

## Acknowledgements

We gratefully acknowledge the support of NVIDIA Corporation with the donation of the Titan Xp GPU used for this research. We would also like to thank the authors of the Deep Image Matting [33] paper for providing us with their training dataset and their evaluation code. Finally, this publication has emanated from research conducted with the financial support of Science Foundation Ireland (SFI) under the Grant Number 15/RP/2776.

# 6 Supplementary material

## 6.1 Architecture of the proposed generator

| Encoder | | | Decoder | | |
|---|---|---|---|---|---|
| layer name | output size | filter size | layer name | output | filter size |
| conv1 | 160×160 | 7×7,64, stride 2 | bilinear | 80×80 | bilinear upsampling |
| conv2_x | 80×80 | 3×3 max pool, stride 2 <br> [1×1, 64; 3×3, 64; 1×1, 256] ×3 | deconv1_x | 80×80 | skip from conv2_x, 1×1,48 <br> [3×3, 256; 3×3, 128; 3×3, 64] |
| conv3_x | 40×40 | [1×1, 128; 3×3, 128; 1×1, 512] ×4 | unpooling | 160×160 | 2×2 unpool, stride 2 |
| conv4_x | 40×40 | [1×1, 256; 3×3, 256 $r=2$; 1×1, 1024] ×6 | deconv2_x | 320×320 | skip from conv1_x, 1×1,32 <br> [3×3, 64; 3×3, 64, stride$\frac{1}{2}$; 3×3, 32] |
| conv5_x | 40×40 | [1×1, 512; 3×3, 512 $r=4$; 1×1, 2048] ×3 | deconv3_x | 320×320 | skip from RGB image <br> [3×3, 32; 3×3, 32] |
| aspp | 40×40 | 1×1, 256; 3×3,$r=6$, 256; 3×3,$r=12$, 256; 3×3,$r=18$, 256; Image Pooling, 256 | deconv4_x | 320×320 | 3×3, 1 |

Table 4: Architecture of the proposed generator. The encoder consists of the standard Resnet50 architecture with the last two layers removed and ASPP [3] module added to output 256 40 × 40 feature maps. The decoder is kept small and uses bilinear interpolation, unpooling and fractionally-strided convolution to upsample the feature maps back to 320 × 320. For the max-pooling operation in the encoder, the maximum indices are saved and used in the unpooling layer. All convolutional layers except the last one are followed by batch-normalization layers [17] and ReLU activation functions. The last convolutional layer is followed by a sigmoid activation function to scale the output between 0 and 1. $r$ is the dilation rate of the convolution. The default stride or dilation rate is 1. Skip connections are added to retain localized information.

## 6.2 Examples from the Composition-1k dataset

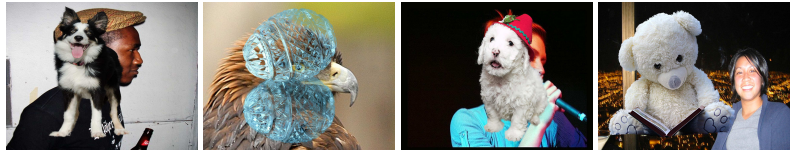

Figure 4: Examples of non-realistic images introduced in the Composition-1k test dataset.

## 6.3 Additional comparison results on the Composition-1k test dataset

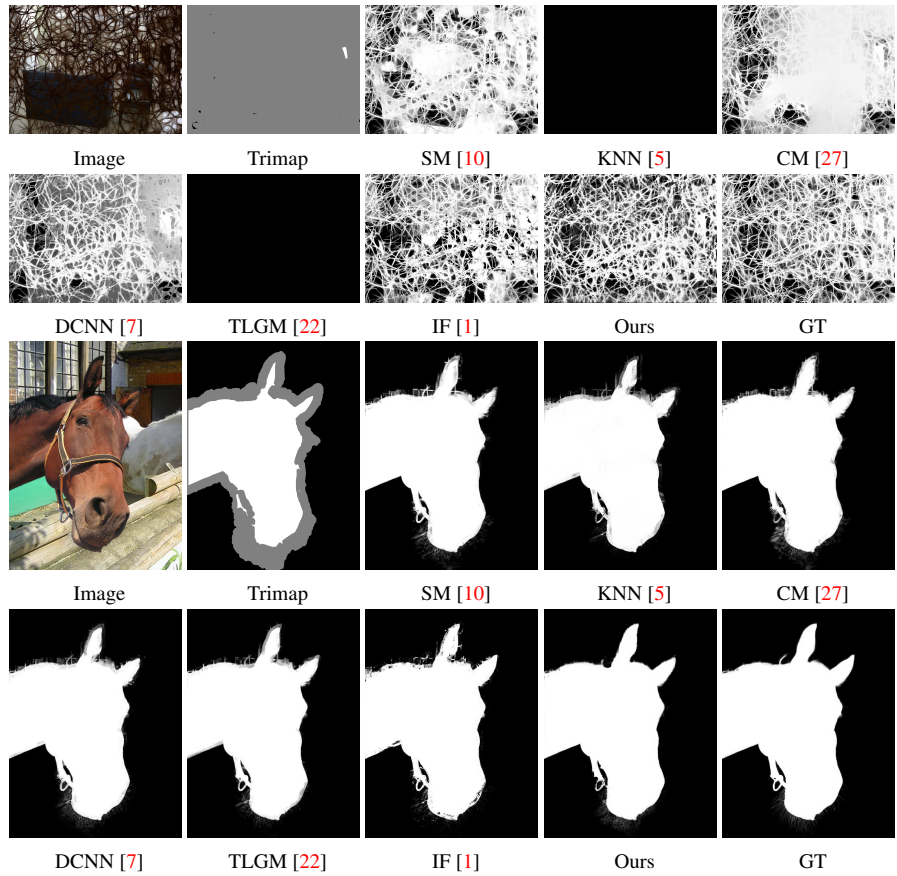

Figure 5: Comparison results on the Composition-1k test dataset.

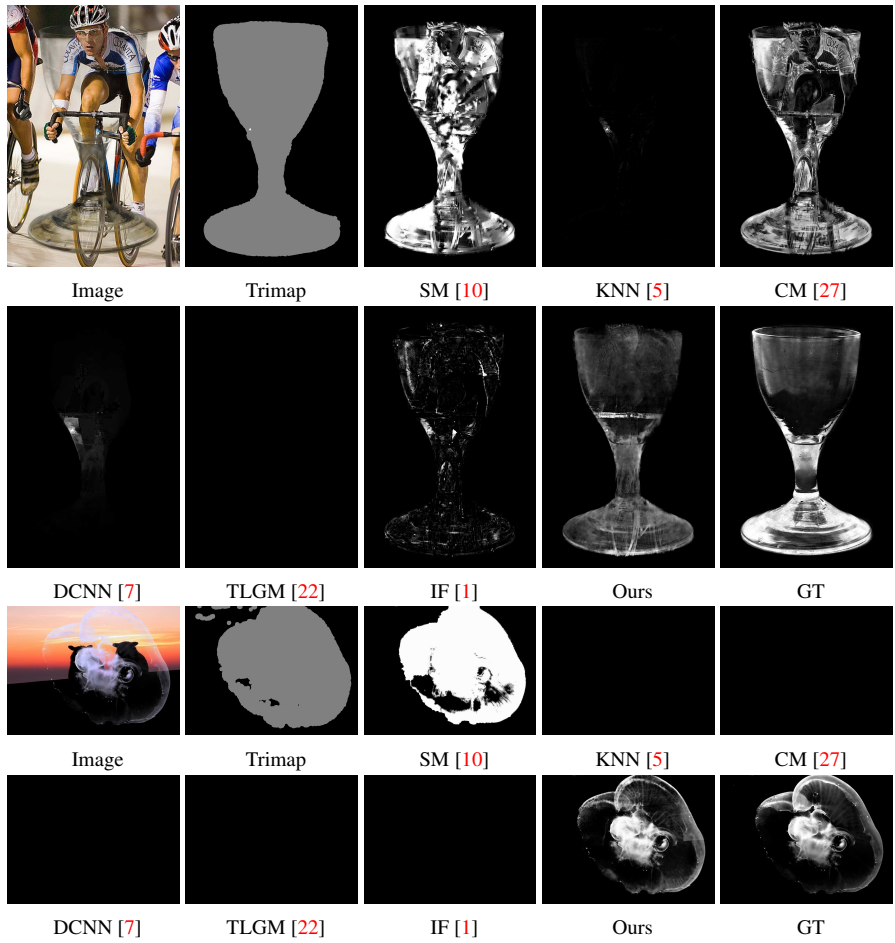

Figure 6: Comparison results on the Composition-1k test dataset.